\documentclass[runningheads]{llncs}

\usepackage{graphicx}
\usepackage{url}
\usepackage{hyperref}
\usepackage{enumitem}
\usepackage{caption}
\usepackage[english]{babel}
\usepackage[utf8]{inputenc}
\usepackage{authblk}
\usepackage{amsmath}
\usepackage{amsfonts}
\usepackage{lipsum}
\usepackage{graphicx}

\usepackage[colorinlistoftodos]{todonotes}
\usepackage{algorithm}
\usepackage{algpseudocode}

\begin{document}

\title{OntoSeer - A Recommendation System to Improve the Quality of Ontologies}
\titlerunning{OntoSeer}

\let\SUP\textsuperscript

%

\author{Pramit Bhattacharyya\inst{1}\orcidID{0000-0002-3611-358X}\thanks{This work was done when the first author was at IIIT-Delhi.} \and Raghava Mutharaju\inst{2}\orcidID{0000-0003-2421-3935}}

\authorrunning{P. Bhattacharyya \and R. Mutharaju}

\institute{Department of Computer Science, IIT Kanpur, Uttar Pradesh, India. \\ \email{pramitb@cse.iitk.ac.in} \and Knowledgeable Computing and Reasoning Lab, IIIT-Delhi, New Delhi, India. \\ \email{raghava.mutharaju@iiitd.ac.in}}


%
\maketitle              
\begin{abstract}

Building an ontology is not only a time-consuming process, but it is also confusing, especially for beginners and the inexperienced. Although ontology developers can take the help of domain experts in building an ontology, they are not readily available in several cases for a variety of reasons. Ontology developers have to grapple with several questions related to the choice of classes, properties, and the axioms that should be included. Apart from this, there are aspects such as modularity and reusability that should be taken care of. From among the thousands of publicly available ontologies and vocabularies in repositories such as Linked Open Vocabularies (LOV) and BioPortal, it is hard to know the terms (classes and properties) that can be reused in the development of an ontology. A similar problem exists in implementing the right set of ontology design patterns (ODPs) from among the several available. Generally, ontology developers make use of their experience in handling these issues, and the inexperienced ones have a hard time. In order to bridge this gap, we propose a tool named OntoSeer, that monitors the ontology development process and provides suggestions in real-time to improve the quality of the ontology under development. It can provide suggestions on the naming conventions to follow, vocabulary to reuse, ODPs to implement, and axioms to be added to the ontology. OntoSeer has been implemented as a Prot\'eg\'e plug-in.  \\

\textbf{Resource Type:} Software tool \\
\textbf{Code Repository:} \url{https://github.com/kracr/ontoseer} \\
\textbf{Demo Video URL:} \url{https://www.youtube.com/watch?v=LGXHGXmVanI}

\end{abstract}

\section{Introduction}

An ontology is a shared understanding of some domain of interest~\cite{Guarino2009}. Ontology development is generally a group activity, where domain experts, ontology developers, and other stakeholders meet to discuss and develop an ontology. This makes ontology development time consuming and expensive, especially if the scope of the ontology is broad. It is not always possible to take the help of domain experts due to a variety of reasons such as their unavailability and cost. Experienced developers face issues while building ontologies, and this problem only magnifies in the case of inexperienced ontology developers. They will have to deal with several questions such as the classes and properties to reuse, the appropriate class hierarchy to build, the axioms to use and the ODPs to use in order to make the ontology modular.

Experienced ontology developers may be able to answer some of these questions and make better design decisions. But even for them, it will be hard to keep track of all the new vocabularies, ontologies, and the ontology design patterns~\cite{567} that get published in the repositories. Our tool, named \emph{OntoSeer}~\cite{10.1145/3430984.3431067,M.TechThesis}, is a Prot\'eg\'e plug-in that works with the ontology developer during the ontology development process and gives suggestions in real-time that can lead to better quality ontologies. The contributions of this work are as follows.
\begin{itemize}
    \item A tool that recommends classes, properties, axioms, and the ontology design patterns to (re)use based on the description, competency questions, and the ontology under development.
    \item A mechanism to provide suggestions of names to use during the ontology development. These suggestions follow the naming conventions.   
    \item Integration with Prot\'eg\'e (available as a plug-in) to improve the ontology development experience.
\end{itemize}



\section{Approach}

In this section, we discuss the various features of OntoSeer, along with the implementation details for each of those features. Since OntoSeer makes use of the existing ontologies and vocabularies in making the recommendations, we first describe the datasets that are used, and then proceed with the description of the method used to make the recommendations. OntoSeer uses the following datasets.


\begin{enumerate}[label=\alph*)]
    \item \textbf{Competency questions (CQs)}. 92 CQs and their corresponding ontologies are available at Software Ontology\footnote{\url{https://softwareontology.wordpress.com/2011/04/01/user-sourced-competency-questions-for-software}}. 52 CQs and ontologies are available at ArCo\footnote{\url{https://github.com/ICCD-MiBACT/ArCo}}. Several CQs and their associated ontologies are available from CORAL~\cite{hitzler_coral:_2019} and~\cite{ren_towards_2014} .
    
    \item \textbf{Ontologies}. We have collected ontologies from several repositories such as NCBO BioPortal\footnote{\url{https://bioportal.bioontology.org/ontologies}}, Manchester OWL Corpus\footnote{\url{http://mowlrepo.cs.manchester.ac.uk/datasets/mowlcorp/}}, Oxford OWL Repository\footnote{\url{https://www.cs.ox.ac.uk/isg/ontologies/}}, and Prot\'eg\'e Ontology Library\footnote{\url{https://protegewiki.stanford.edu/wiki/Protege\_Ontology\_Library}}. 
    
    \item \textbf{Ontology Design Patterns (ODPs)}. ODPs are available at the ODP repository\footnote{\url{http://ontologydesignpatterns.org/wiki/Main\_Page}}. Since there is no bulk download option, we used a web scraper to collect all the ODPs. There are six categories of ODPs, and their URLs are used as the seed for the web scraper.
    
    \item \textbf{Vocabularies}. Several vocabularies are indexed at the Linked Open Vocabularies (LOV)\footnote{\url{https://lov.linkeddata.es/dataset/lov/}}. These vocabularies can be accessed using either the SPARQL endpoint or the LOV API. 
\end{enumerate}

\subsection{Class, Property and Vocabulary Recommendation}

One of the recommended best practices in building an ontology is to reuse existing ontologies. But there are several ontologies spread across many repositories which makes it hard to find the right ones to reuse. OntoSeer creates an inverted index of the ontologies from the repositories based on the classes and properties found in the ontologies. It queries this index to get a ranked list of the classes and properties to reuse, along with the ontology that they are part of. These classes and properties are similar to the ones present in the ontology that is currently being built. In the case of repositories that can be queried over the Web, such as LOV\footnote{\url{https://lov.linkeddata.es/dataset/lov/api/v2/term/suggest?q=}} and BioPortal\footnote{\url{http://data.bioontology.org/recommender?input=}}, we do so without making a local copy (index).

\begin{figure}
\centering
\includegraphics[width=8.5cm, height=3.5cm]{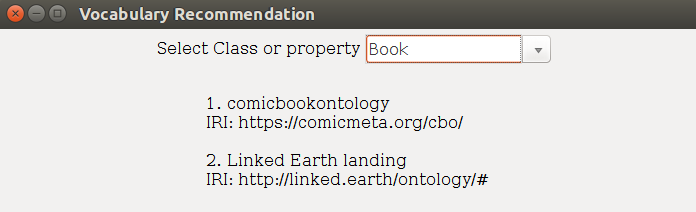}
\caption{Vocabulary recommendation from OntoSeer for the class \texttt{Book}}
\label{fig:HI}
\end{figure}

Figure~\ref{fig:HI} shows OntoSeer's recommendations for a class \texttt{Book}. The first recommendation is of \emph{Comic Book Ontology} along with its IRI. The second description represents the vocabulary \emph{Linked Earth Landing} along with its IRI. In both these vocabularies, a class named \texttt{Book} is present and the user can choose to reuse this class from one of the two vocabularies instead of creating a new one.

If the competency questions (CQs) are provided by the user, OntoSeer recommends the classes and properties to start with by identifying the nouns and the verbs in the CQs.

\subsection{Class and Property Name Recommendation}

Following the appropriate naming conventions for classes and properties helps in improving the readability and the ease of maintenance of an ontology. But it is a challenge to abide by the naming conventions, especially for a beginner, because they may not be familiar with the conventions. To cater to this need, OntoSeer supports the class and property naming recommendations available at \url{http://www-sop.inria.fr/acacia/personnel/phmartin/RDF/conventions.html}. They can be summarized as follows.

\begin{itemize}
    \item Use of numbers in the name is discouraged. If there is a class named \texttt{Human1234} in the ontology, a suggestion to use \texttt{Human} is given.
    \item Use of any special character other than an underscore is not recommended.
    \item Use of upper camel case while naming is encouraged. If a class is named \texttt{Human\_being}, the recommendation would be to use \texttt{HumanBeing} instead. 
\end{itemize}   

\begin{figure}[ht]
 \centering
\includegraphics[width=8cm, height=4cm]{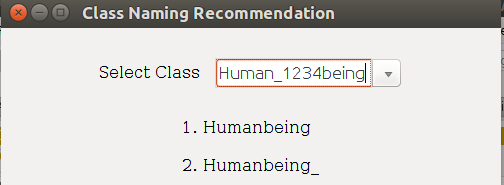}
\caption{OntoSeer removed the digits present in the classname and suggested names that follow the upper camel case convention.}
\label{fig:I}
\end{figure}

A screenshot of the naming recommendation is shown in Figure~\ref{fig:I}.

\subsection{Axiom Recommendation}

\begin{figure}[ht]
 \centering
\includegraphics[width=7cm, height=4cm]{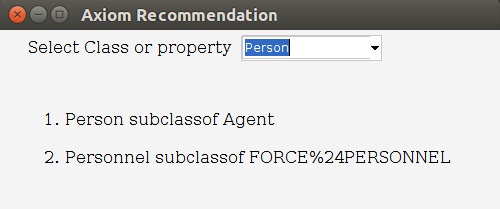}
\caption{Axiom recommendation for the class \texttt{Person}}
\label{fig:Gii}
\end{figure}

OntoSeer can retrieve axioms from the ontology corpus that have the closest match with the entities (classes, properties) of the current ontology. It will recommend axioms such as disjointedness among class siblings and property characteristics (inverse, symmetric, transitive, etc.). The inverted index structure created on the ontology corpus is used to make the recommendations. A sample recommendation of axioms involving the class \texttt{Person} is shown in Figure~\ref{fig:Gii}. The top two or three recommendations are shown based on the similarity score. The similarity threshold has been varied between 0 to 1 with an increment of 0.05 and has been fixed at 0.85 after observing the recommendations on 150 different instances.


\subsection{ODP Recommendation}

Ontology Design Patterns help in making an ontology modular and also reduce the effort in building an ontology since the design solutions approved by the community can be reused. But it is hard to find the right ODP from among the hundreds of ODPs available at the ODP repository\footnote{\url{http://ontologydesignpatterns.org/wiki/Main\_Page}}. OntoSeer helps in easing this process by recommending appropriate ODPs for a given ontology by considering its classes and properties. Apart from the classes and properties of an ontology, OntoSeer can also take into account any available description, competency questions and the domain of the ontology while making the recommendations. All the ODPs available at the ODP repository have been collected. Each ODP has several fields associated with it, such as, intent, domain, competency questions, elements and the description.



OntoSeer makes use of string similarity for ODP recommendation. We compare the classes and the properties with those from the ODPs. These are considered as two sets of strings. Jaro-Winkler~\cite{unknown} distance, a variant of edit distance is used for computing the similarity between these two sets of strings. We have tried to use  Doc2Vec~\cite{DBLP:conf/icml/LeM14} because it considers semantic matching. The incompatibility of deeplearning4j\footnote{\url{https://deeplearning4j.org/}} with Prot\'eg\'e compelled us to use the Jaro-Winkler method, and this gives comparable results.
Jaro-Winkler method is able to mimic the results obtained from Doc2Vec to a considerable extent, as shown in~\cite{Cahyono_2019}.






We gave more weightage to the similarity scores obtained from class and property matching followed by the scores for the  description of the ontology, domain of the ontology, and competency questions. The total score, thus, can be represented as   $ODP_{score}$ = $w_1*s_1$ + $w_2*s_2$ + $w_3*s_3$ + $w_4*s_4$, where $s_1$ is the similarity score obtained from term (class and property) matching, $s_2$ is the score obtained from matching the description of the ontology, $s_3$ is the score obtained from matching the domain of the ontology, and $s_4$ is the score obtained from matching the competency questions. The weightage for components ($w_1$=5, $w_2$=3, $w_3$=2, $w_4$=3) and a similarity threshold of 0.65 has been fixed after running the experiments for more than 150 iterations.

\begin{figure}[ht]
 \centering 
\includegraphics[width=12cm, height=6cm]{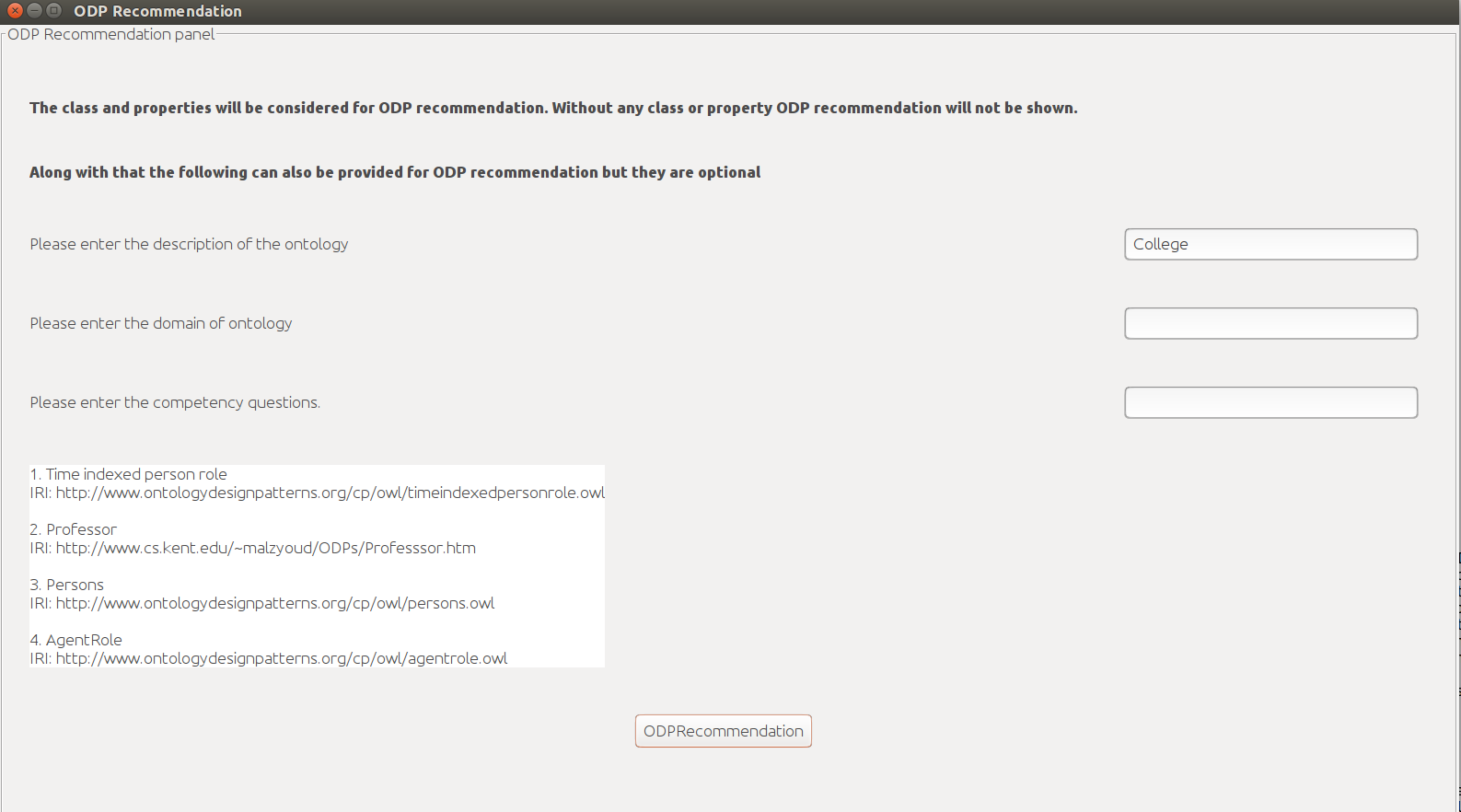}
\caption{OntoSeer's ODP recommendations on providing \emph{College} as the description and \texttt{Person}, \texttt{Professor} are the classes available in the ontology}
\label{fig:G}
\end{figure}

If an ontology has classes named \texttt{Person} and \texttt{Professor} and the description provided for the ontology is \emph{College}, the recommended ODPs are \emph{TimeIndexedPersonRole},  \emph{Professor}, \emph{Persons} and \emph{AgentRole}. This is shown in Figure~\ref{fig:G}. On analyzing the recommendations, we noticed that these ODPs and their descriptions have the same or similar terms (classes and properties) as the ones found in the ontology. 

\subsection{Class Hierarchy Validation}
Determining the class hierarchy is confusing, especially to the inexperienced ontology developers. This often leads to improper hierarchy. OntoSeer makes use of rigidity (R), identity (I) and unity (U) characteristics~\cite{guarino_overview_2009} of the classes and takes the user input based on these characteristics to validate the class hierarchy. These three characteristics can be defined as follows.
\begin{enumerate}
  \item \textbf{Rigidity:} A property is rigid if it is essential to
all its possible instances. An instance of a rigid property cannot cease to exist in the future or fail to be its instance in a different domain. 
\item \textbf{Identity:} A property carries an identity criterion  if all its instances can be identified by means of a suitable \emph{sameness} relation. Identity constraint is the criteria we use to answer questions like, ``Is that my dog?''.
\item \textbf{Unity:} A property P is said to carry unity if there is a common unifying relation R such that all the instances of P are wholes under R. Unity constraint checks whether properties have wholes as instances.  
\end{enumerate}

OntoSeer interacts with the ontology developer and based on the inputs, determines the characteristics of a particular class. The questions asked by OntoSeer are as follows.
\begin{enumerate}
    \item  Do the properties of the class cease to exist in the future? For example, \texttt{Person} will always be a person, but \texttt{Student} can cease to exist to be a student after a time lapse.
    \item Are the properties of superclass and subclass identical? For example, two one-hour duration time intervals are identical, but an hour interval on Wednesday is not identical to an hour interval on Friday.
    \item Is the property of the subclass part of the properties of the super class? For example, a lump of clay is part of ``amount of matter'', but ``amount of matter'' is not part of a lump of clay.
\end{enumerate}

\begin{figure}[ht]
 \centering 
\includegraphics[width=12cm, height=6cm]{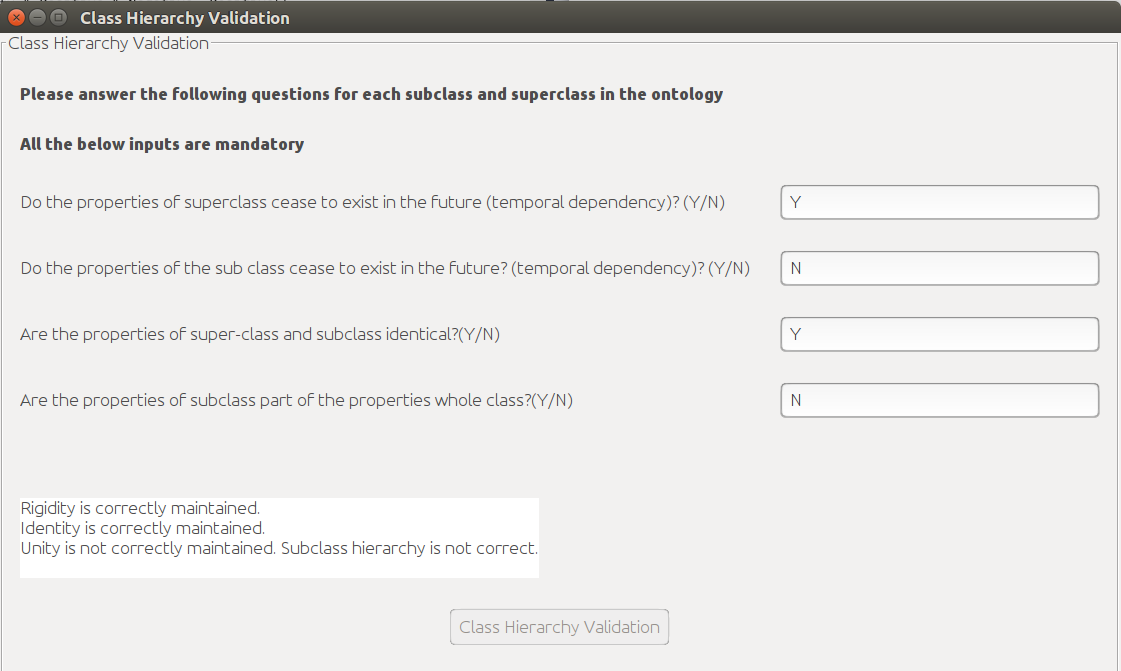}
\caption{OntoSeer validates the class hierarchy based on the  rigidity, identity and unity characteristics. Here, the unity constraint is violated and hence the class hierarchy needs to be revisited.}
\label{fig:class-hierarchy}
\end{figure}

Table~\ref{tab:Table1} shows the superclass and its accepted subclass characteristics for validating the class hierarchy. Consider the first row of the table. If the superclass is satisfying the Identity criteria, the Identity constraint for the hierarchy will be maintained only when the subclass also satisfies the Identity criteria. The rest of the rows of Table~\ref{tab:Table1} can be interpreted in the same way. A screenshot from OntoSeer validating the class hierarchy based on the class characteristics is given in Figure~\ref{fig:class-hierarchy}.

\begin{table}[ht] 
\begin{center}
 \caption{The values of rigidity, identity and unity for class hierarchy validation.}
\label{tab:Table1}
\begin{tabular}{ ccc } 
 \hline
  Rule & SuperClass Value & SubClass Value \\ [0.05ex] 
 \hline\hline
 Identity & Positive & Positive\\ 
 Identity & Negative &Negative\\ 
 Rigidity & Positive & Negative, Positive\\ 
 Rigidity & Negative &Negative\\ 
 Unity & Positive &  Positive\\ 
 Unity & Negative &Negative\\ 
 \hline
\end{tabular}
\end{center}
\end{table}

\section{User Study}
We evaluated OntoSeer through user study by considering the following questions.
\begin{enumerate}
    \item Is OntoSeer useful for the inexperienced as well as the experienced ontology developer? 
    \item Is OntoSeer saving the time of ontology developers?
    \item Are ontology developers able to make richer ontologies due to the axiom recommendations of OntoSeer?
    \item Are the ontology developers reusing existing ontologies more than they generally do?
    \item How easy is it to install and use OntoSeer? 
    \item Are the ontology developers able to create modular ontologies by incorporating ODPs?
    \item Are the ontology developers able to build ontologies that follow the naming conventions?
\end{enumerate}


We advertised OntoSeer across several relevant mailing lists\footnote{Prot\'eg\'e users, ODP users, Semantic Web, OWL API} and contacted some of the research groups that work on ontology design. We requested them to use OntoSeer and provide an anonymous response to a questionnaire\footnote{\url{https://github.com/kracr/ontoseer/blob/master/Evaluation/User\%20Study/User\%20Study\%20Questions.txt}}. It had questions on the modelling experience of the users, their familiarity with ODPs, the ease of installing OntoSeer, the relevance of the recommendations provided by OntoSeer, the modelling experience with and without using OntoSeer and whether OntoSeer helps in saving modelling time.

There were 21 respondents. Of the 21 users, 12 have identified themselves as having basic proficiency in Prot\'eg\'e, 7 are at intermediate level and 2 have advanced proficiency in ontology modelling. For evaluating OntoSeer, users can either build an ontology or import an existing ontology. 



12 users said that the plugin's installation is easy, and the instructions were clear, whereas 7 users found it moderately difficult to install the plugin. Table~\ref{tab:Table5} represents the cumulative user response for vocabulary, ODP and axiom recommendation when the CQs are provided. Table~\ref{tab:Table6} represents the cumulative user response for the same features without the CQs. The ratings are on a scale of 1 to 5, with 1 being the lowest (not satisfied) and 5 is the highest (very satisfied).

\begin{table}[ht] 
\begin{center}
\caption{Average satisfactory index of users for OntoSeer features without CQs}
\label{tab:Table6}
\begin{tabular}{ cccccc } 
 \hline
 Proficiency & Vocabulary & ODP & Axiom  \\ 
 Type  & Recommendation & Recommendation & Recommendation \\
 \hline\hline
 Basic & 4 & 3.67 &4\\ 
 Intermediate & 3.571 &3.571 &3.714\\ 
 Advanced& 3 &3.5 &3\\
 \hline
\end{tabular}
\end{center}
\end{table}

\begin{figure}[ht]
\centering
  \begin{tabular}{cc}
    \includegraphics[width=0.5\textwidth]{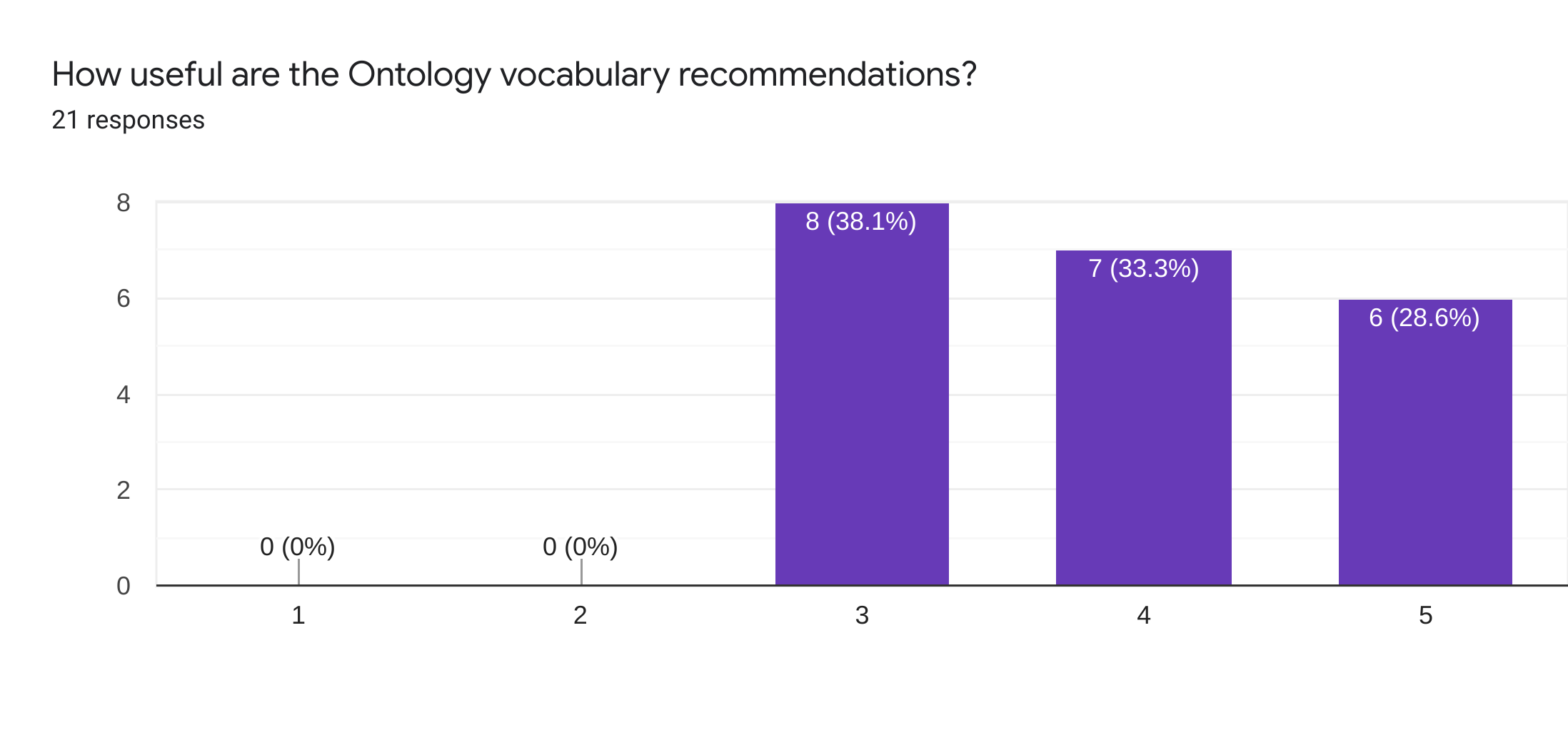} &
    \includegraphics[width=0.5\textwidth]{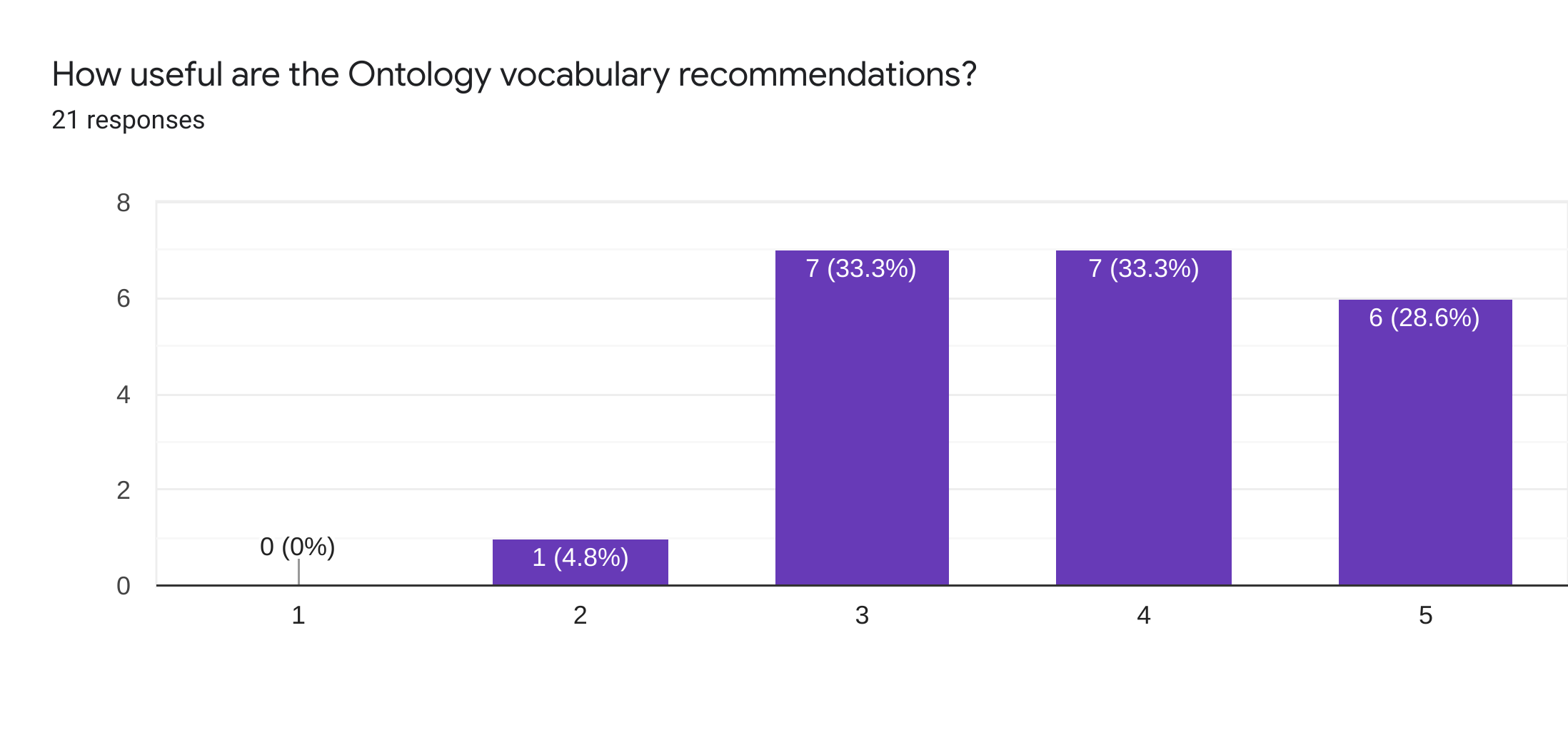} 
\end{tabular}
  \caption{User response on the vocabulary recommendation with and without CQs}
  \label{fig:tableA}
\end{figure}

\begin{figure}[htb]
\centering
  \begin{tabular}{cc}
    \includegraphics[width=0.5\textwidth]{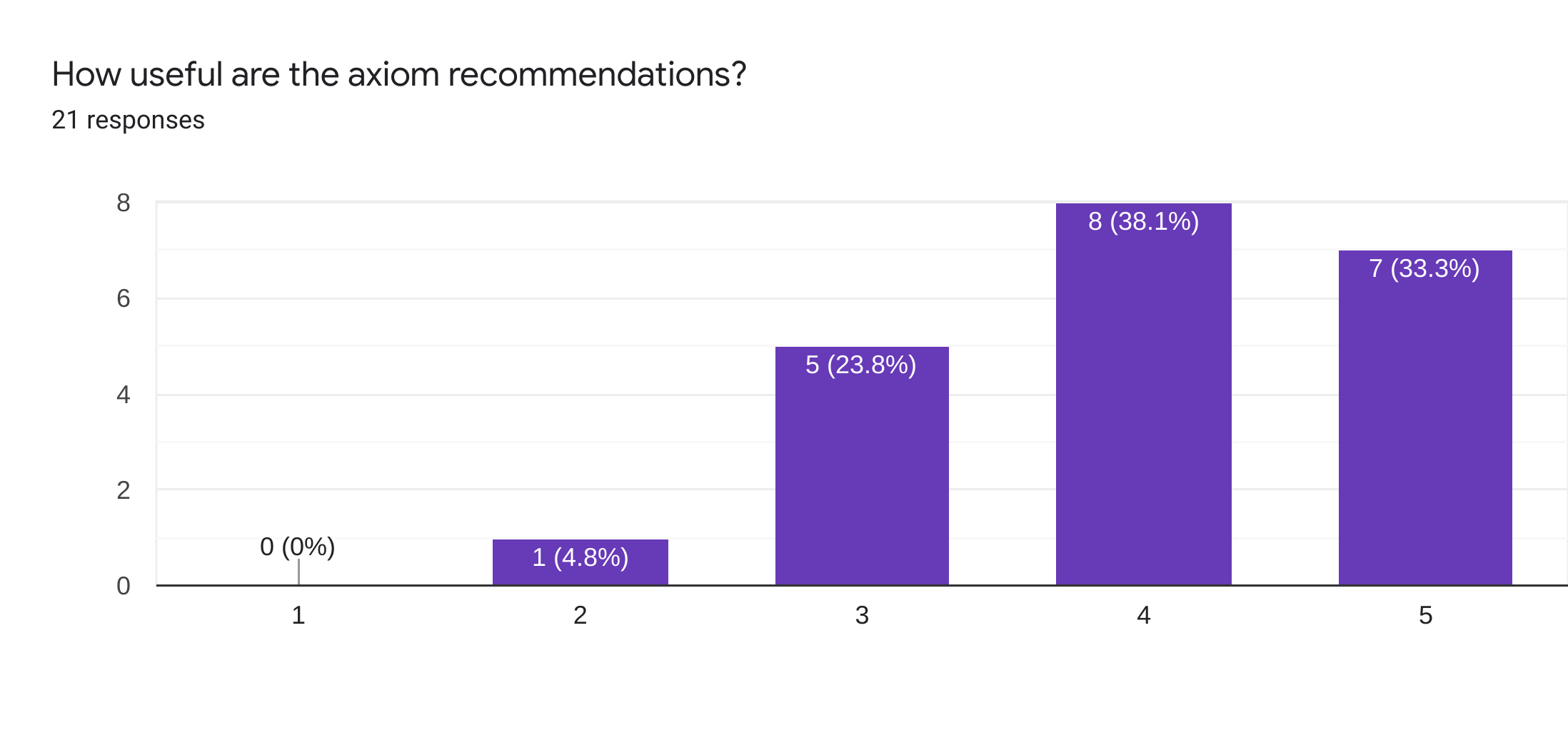} &
    \includegraphics[width=0.5\textwidth]{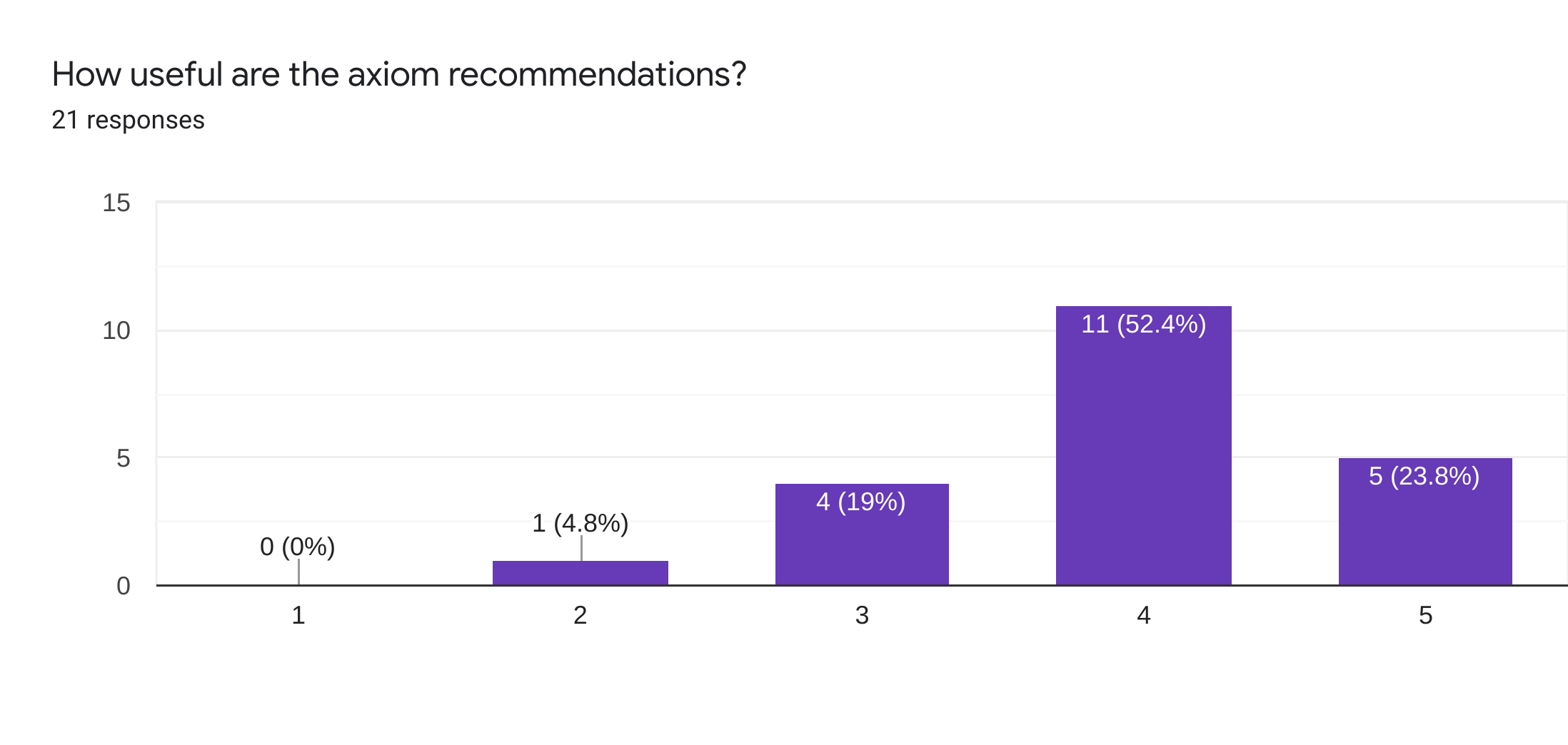} 
 \end{tabular}
  \caption{User response on the axiom recommendation with and without CQs}
  \label{fig:tableB}
\end{figure}

\begin{figure}[htb]
\centering
  \begin{tabular}{cc}
    \includegraphics[width=0.5\textwidth]{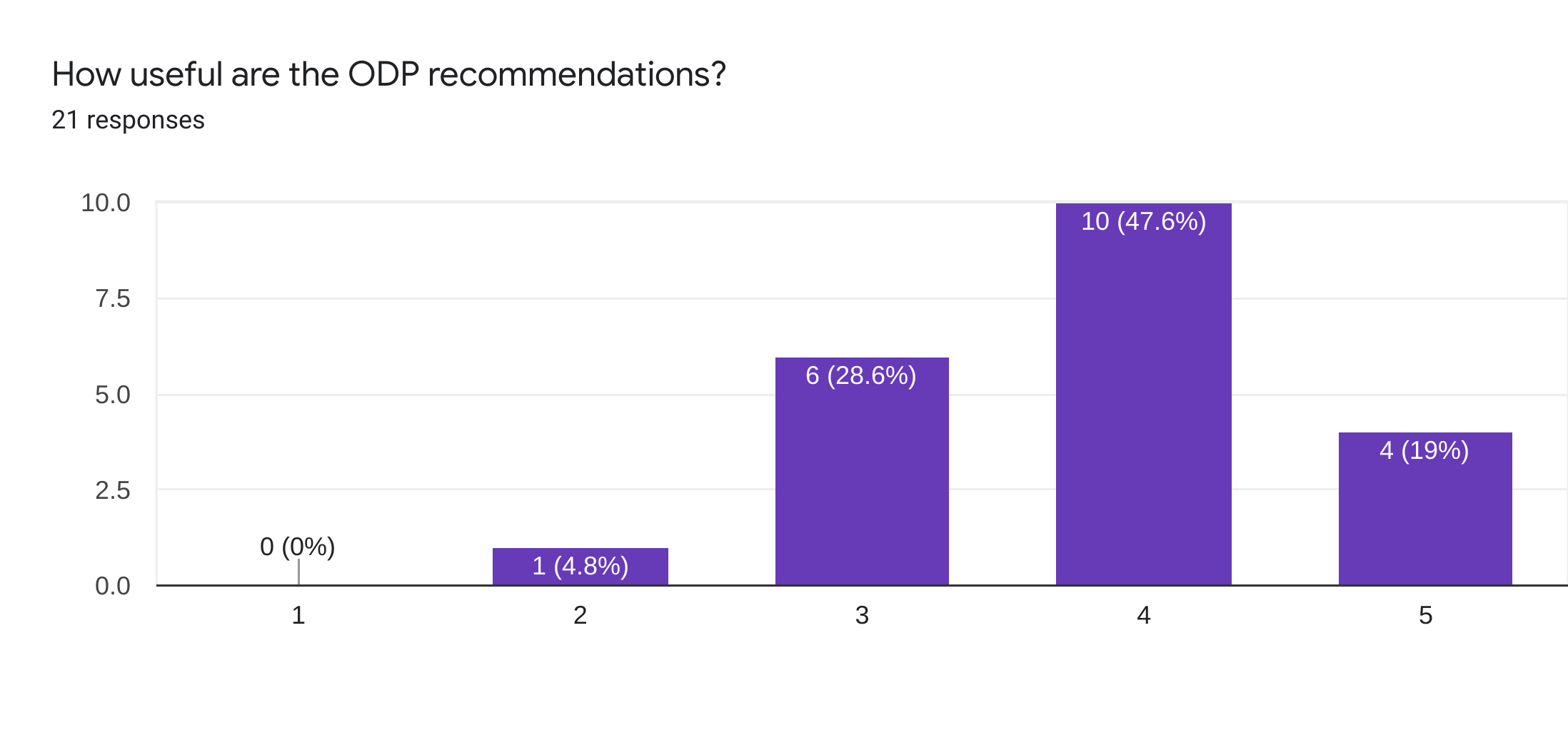} &
    \includegraphics[width=0.5\textwidth]{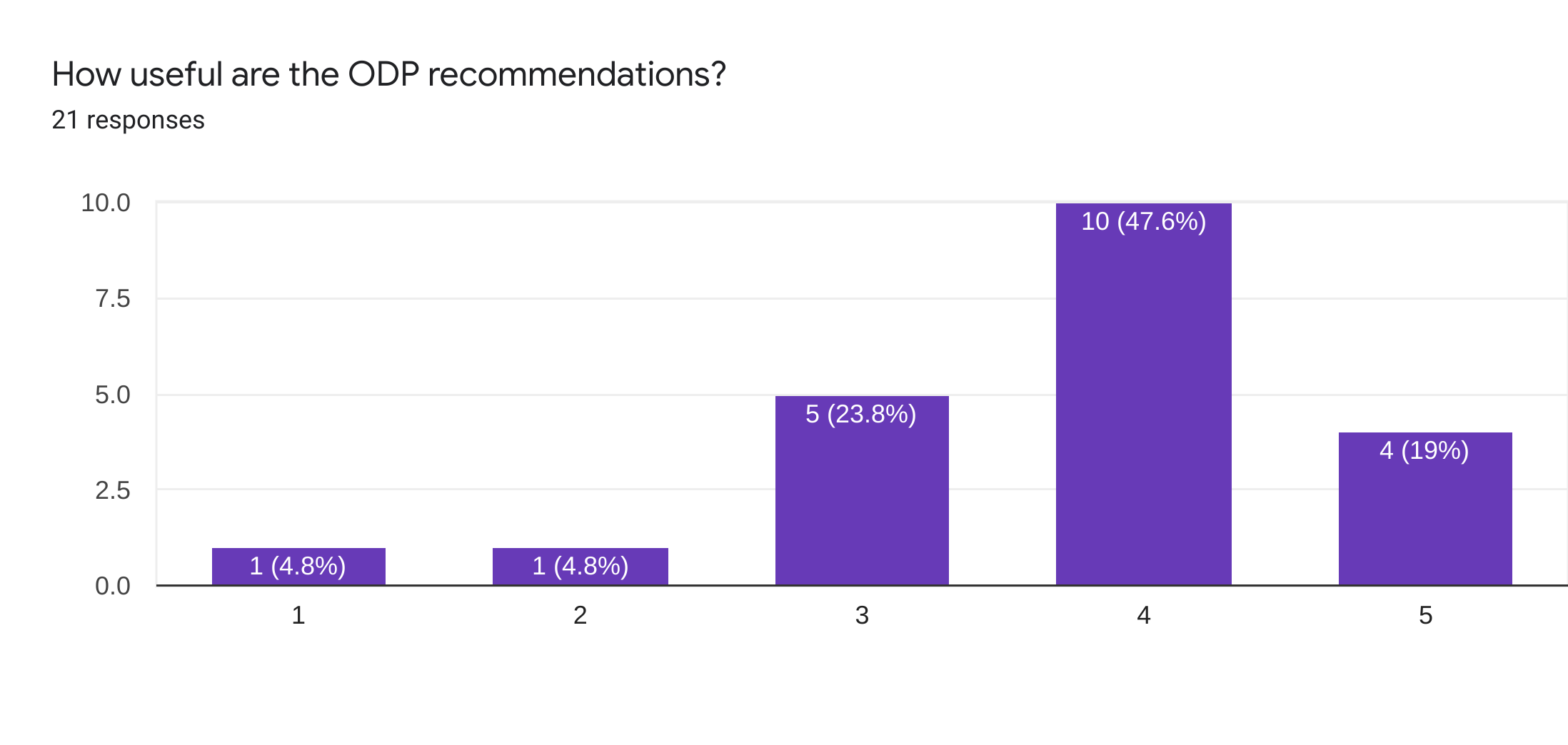} 
 \end{tabular}
  \caption{User response on the ODP recommendation with and without CQs}
  \label{fig:tableD}
\end{figure}

\begin{figure}[htb]

 \centering 
\includegraphics[width=10cm, height=5cm]{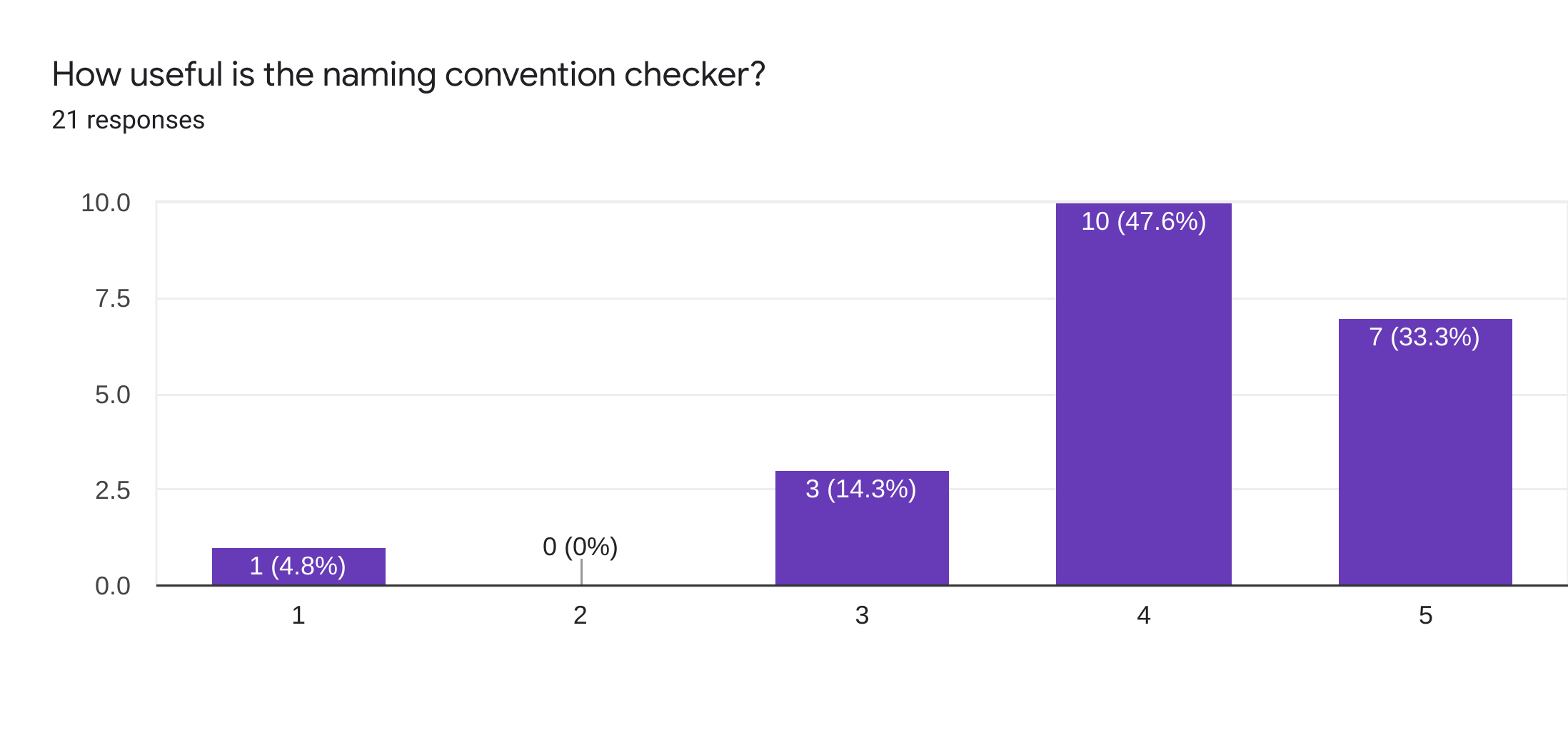}
\caption{User response on the name recommendation}
\label{fig:Naming}

\end{figure}

\begin{figure}[htb]

 \centering 
\includegraphics[width=10cm, height=5cm]{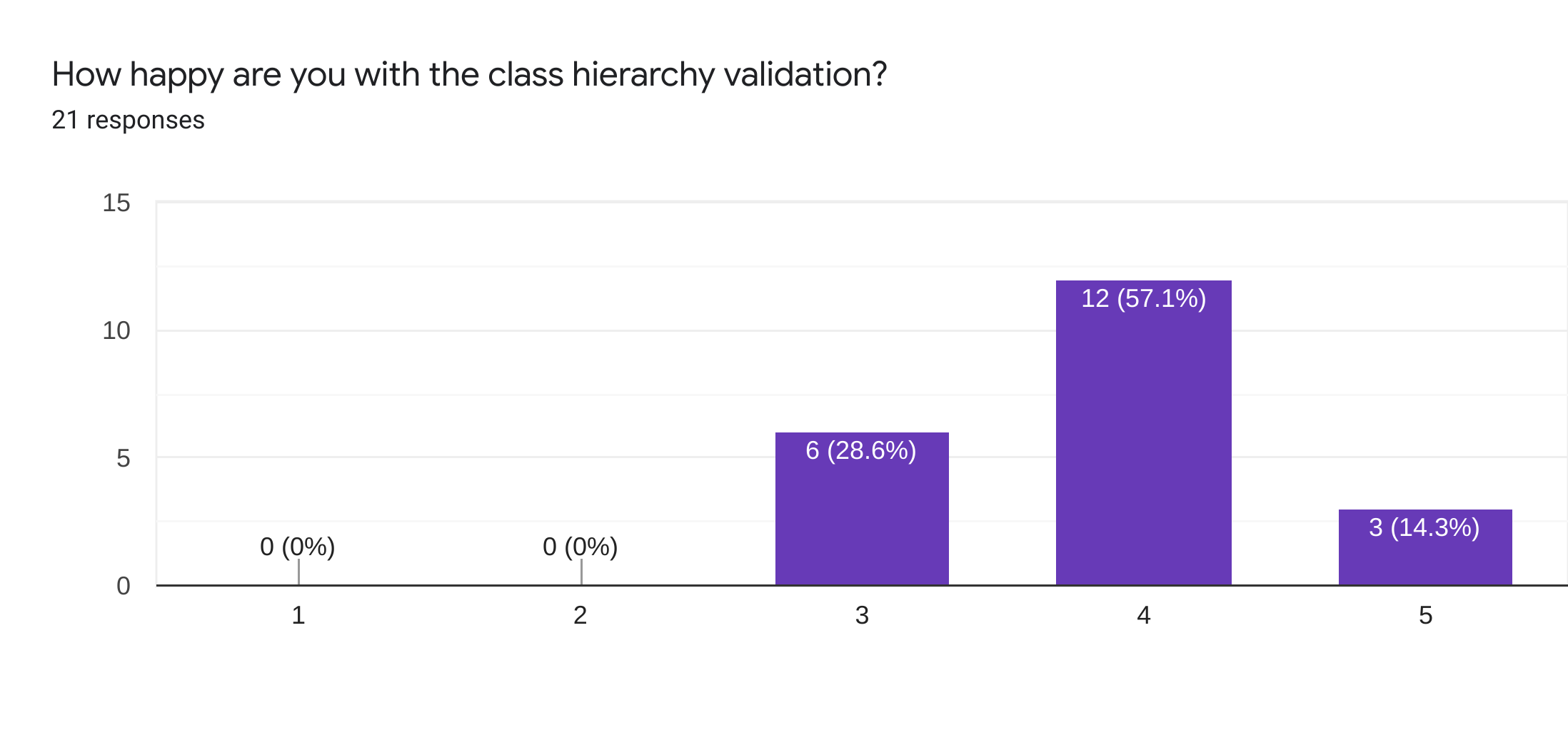}
\caption{User response on the class hierarchy validation}
\label{fig:ClassHierarchy}
\end{figure}

Figures~\ref{fig:tableA} to \ref{fig:ClassHierarchy} are the graphical representations of the user response for the different features of OntoSeer. The average user satisfaction index for name recommendation and class hierarchy validation is given in Table~\ref{tab:Table7}. On an average, users are happy with Ontoseer's recommendations.  Table~\ref{tab:Table5} and Table~\ref{tab:Table6} show that the average satisfaction of users decreases minimally when the intermediate user response is compared with the basic user response (advanced user response is not considered significant since there were only two users). With more modelling experience, users become aware of LOV and ODPs. But even for experienced modellers, OntoSeer can provide useful recommendations. Table~\ref{tab:Table5} and Table~\ref{tab:Table6} also show that the recommendations are better when CQs are available which is an expected result. But the performance of OntoSeer does not drop drastically in the absence of CQs.


\begin{table}[ht] 
\begin{center}
\caption{Average satisfactory index of users for OntoSeer features with CQs}
\label{tab:Table5}
\begin{tabular}{ cccc} 
 \hline
 Proficiency & Vocabulary & ODP & Axiom  \\ [0.1ex]
 Type  & Recommendation & Recommendation & Recommendation \\ [0.1ex]
 \hline\hline
 Basic & 4.25 & 4 &4.083\\ 
 Intermediate & 3.86 &3.86 &4.143\\ 
 Advanced& 3 &3.5 &3.5\\
 \hline
\end{tabular}
\end{center}
\end{table}

\begin{table}[ht] 
\begin{center}
 \caption{Average user satisfactory index for OntoSeer's naming recommendation and class hierarchy validation}
\label{tab:Table7}
\begin{tabular}{ ccc } 
 \hline
 Proficiency & Naming  & Class Hierarchy  \\
 Type & Recommendation & Validation \\
 \hline\hline
 Basic & 4.08 & 3.92\\ 
 Intermediate & 4.14 &4\\ 
 Advanced& 4 &3.5 \\
 \hline

\end{tabular}
\end{center}
\end{table}

\begin{figure}[htb]

 \centering 
\includegraphics[width=9cm, height=4cm]{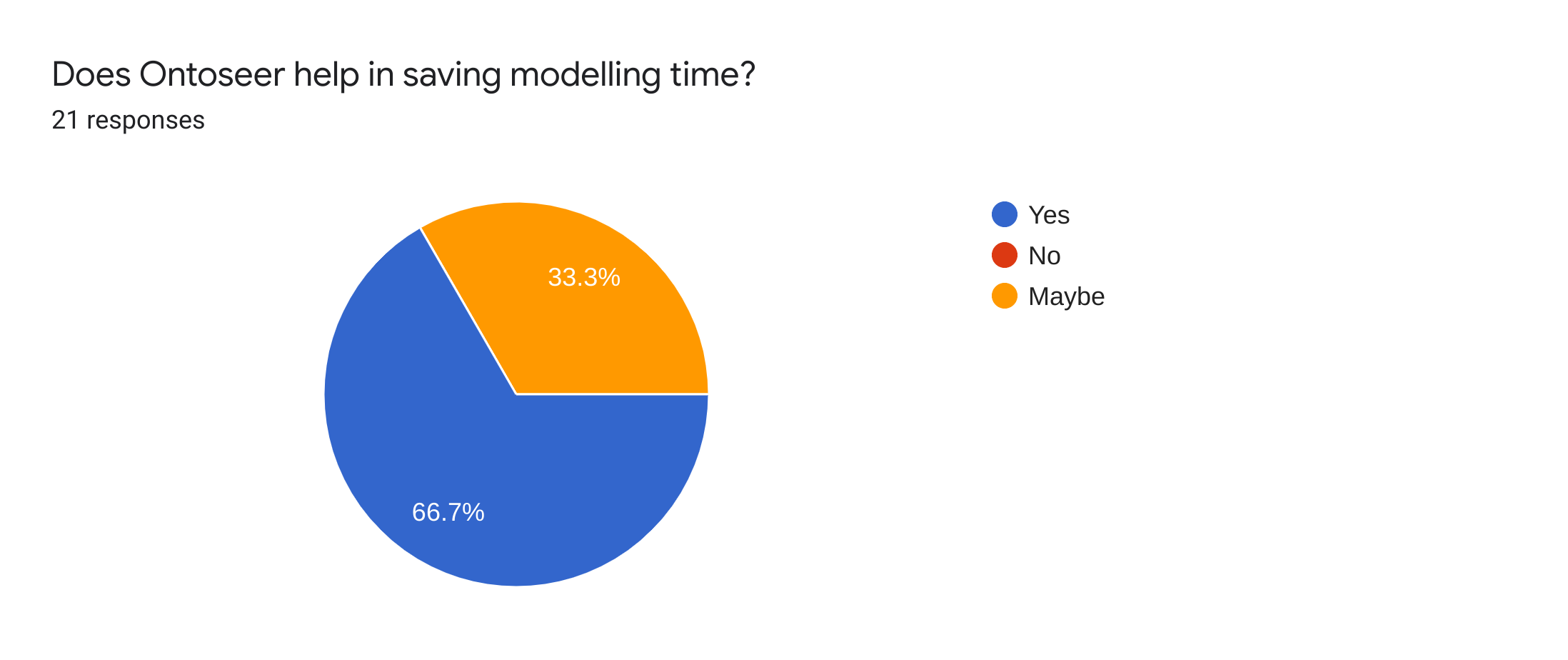}
\caption{User response to the question on whether OntoSeer saves modelling time}
\label{fig:ModellingTime}

\end{figure}

An important aspect in the evaluation of OntoSeer is the quality of ontology that gets developed and the amount of effort in terms of the time taken to develop an ontology. On average, there is an 8.33\% improvement in modelling experience for the basic proficiency level users when using OntoSeer. This increases to 25.72\% for intermediate proficiency level users. The two advanced proficiency level users chose to be neutral. Fourteen, that is, 66.67\% of users believed that OntoSeer saves modelling time while the remaining seven chose to be neutral (Figure~\ref{fig:ModellingTime}). 



\begin{figure}[htb]
\centering
  \begin{tabular}{cc}
  \label{fig:table D}
    \includegraphics[width=0.5\textwidth]{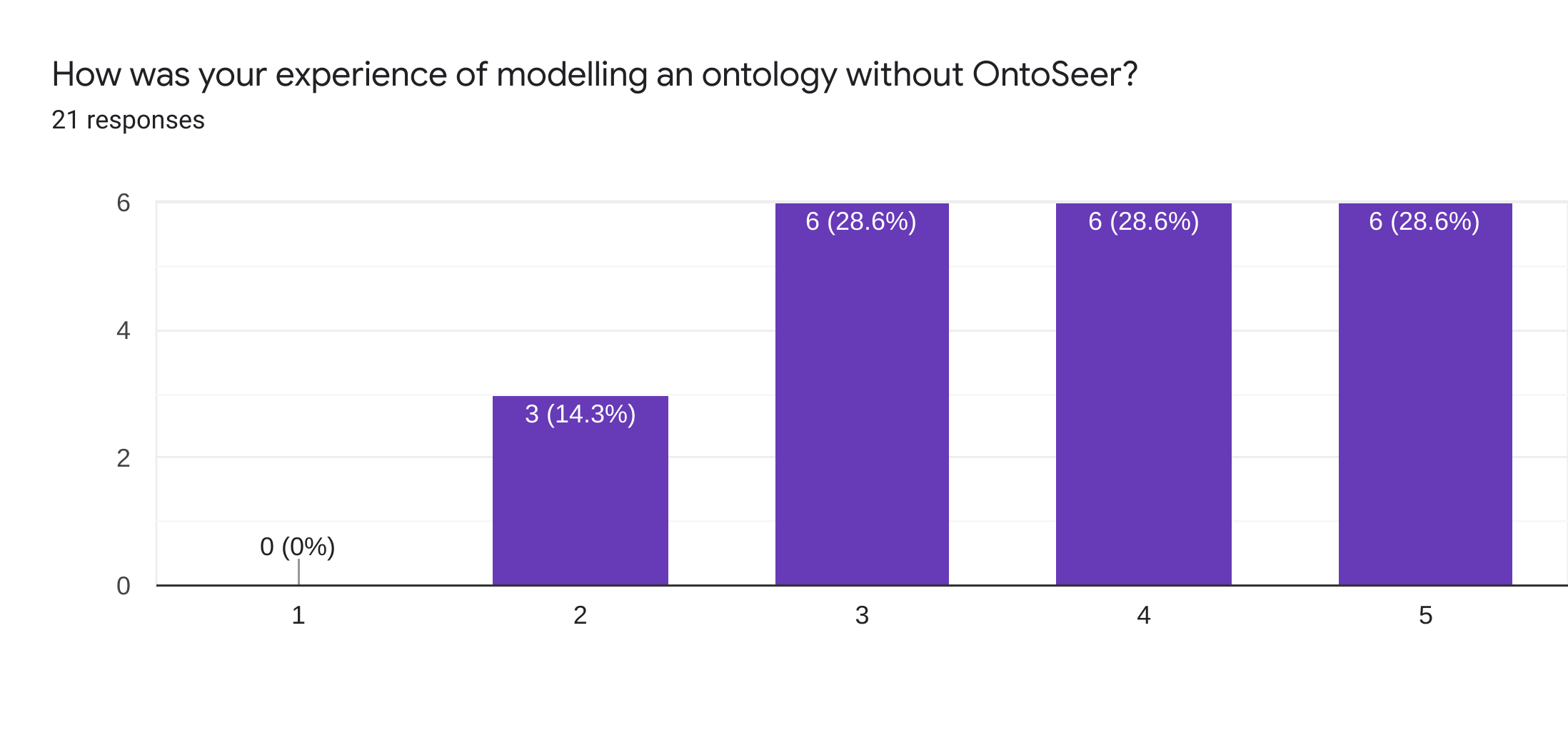} &
    \includegraphics[width=0.5\textwidth]{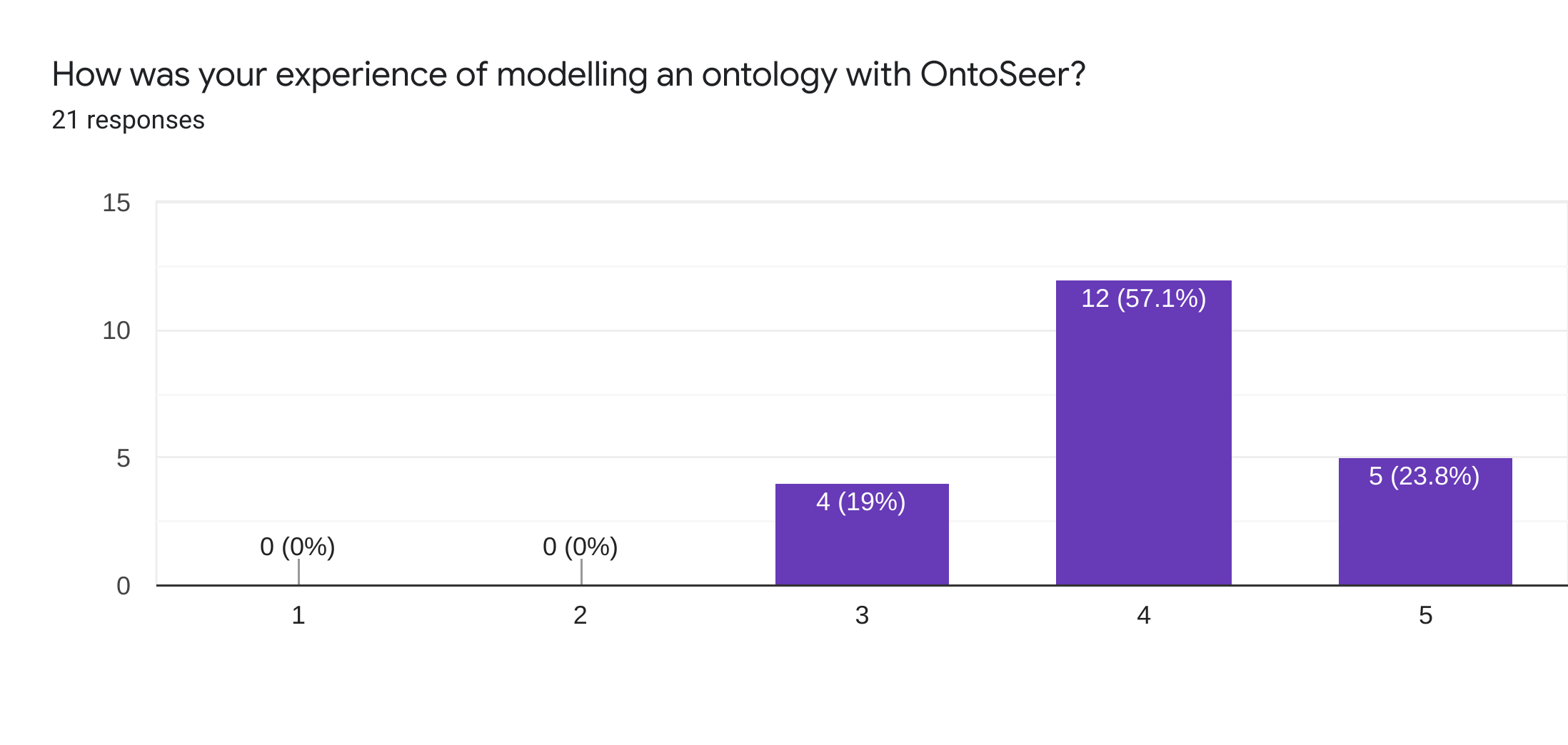} 
\end{tabular}
  \caption{Overall user satisfaction in building ontologies without OntoSeer and with OntoSeer}
\end{figure}

\section{Evaluating the quality of OntoSeer's recommendations}

Using the user study, we evaluated the utility and the effectiveness of OntoSeer's recommendations in real-time, i.e., when it was used as a plug-in while building an ontology in Prot\'eg\'e. There exists the possibility of a bias here because the recommendations were already provided by OntoSeer and the users were asked to evaluate them. In order to avoid this, we evaluate OntoSeer against the users' recommendations, i.e., we first ask the users to make the recommendations which are then compared with OntoSeer's recommendations. We provide two simple use cases in the form of English language descriptions. Along with the descriptions, a few ontologies and ODPs are also provided. Among them, some are relevant to the use case and some are not. Four users were asked to build an ontology (without OntoSeer) by going over each use case and reusing the classes, properties, axioms from the provided ontologies and ODPs. Users worked individually and also in a group of two while building ontologies for the two use cases. The ontologies built by the users, along with the reused classes, properties, axioms from the provided ontologies and ODPs were collected. OntoSeer is then run on the same ontologies and it's recommendations are compared against the ones from the users. In order to mitigate the effects of building the ontologies individually and in a group, we took the union of the recommendations for each use case and compared them with OntoSeer's recommendations. We made these comparisons in two settings. In the exact setting, OntoSeer used only the ontologies and the ODPs provided to the users in order to produce the recommendations. In the general setting, OntoSeer used the entire repository of ontologies and ODPs that were available to it in order to generate the recommendations. Table~\ref{tab:Table17} and Table~\ref{tab:Table18} show the precision and recall of OntoSeer for the use cases 1 and 2 in the general setting. Table~\ref{tab:Table19} and Table~\ref{tab:Table20} show the precision and recall of OntoSeer in the exact setting. From this evaluation, it is evident that OntoSeer's recommendations are relevant and are of good quality. The use cases with all the owl files and user response are available at \url{https://github.com/kracr/ontoseer/tree/master/Evaluation}.

\begin{table}[ht] 
\begin{center}
 \caption{Precision and recall of OntoSeer for use case 1 in the general setting.}
\label{tab:Table17}
\begin{tabular}{ ccccccc } 
 \hline
Feature&Precision@3 & Recall@3 & Precision@5 & Recall@5 & Precison@7& Recall@7\\
 
 \hline\hline
 ODP&0.66&0.4 &0.6&0.6&0.714 &1\\ 
 Vocabularies&0.66&0.33&0.6&0.5&0.714&0.833\\ 
 Axioms&0.33&0.2&0.4&0.4&0.43&0.6\\
 \hline

\end{tabular}
\end{center}
\end{table}

\begin{table}[ht] 
\begin{center}
 \caption{Precision and recall of OntoSeer for use case 2 in the general setting.}
\label{tab:Table18}
\begin{tabular}{ ccccccc } 
 \hline
Feature&Precision@3 & Recall@3 & Precision@5 & Recall@5 & Precison@7& Recall@7\\

 \hline\hline
 ODP&0.33&0.2 &0.6&0.6&0.714 &0.8\\ 
 Vocabularies&0.66&0.4&0.4&0.4&0.571&0.66\\ 
 Axioms&0.33&0.166&0.4&0.33&0.29&0.5\\
 \hline

\end{tabular}
\end{center}
\end{table}

\begin{table}[htb] 
\begin{center}
 \caption{Precision and recall of OntoSeer for use case 1 in the exact setting.}
\label{tab:Table19}
\begin{tabular}{ ccccccc } 
 \hline
Feature&Precision@3 & Recall@3 & Precision@5 & Recall@5 & Precison@7 & Recall@7\\
 
 \hline\hline
 ODP&0.66&0.4 &0.8&0.8&0.714 &1\\ 
 Vocabularies&0.66&0.33&0.66&0.66&0.857&1\\ 
 Axioms&0.66&0.4&0.6&0.6&0.571&0.8\\
 \hline

\end{tabular}
\end{center}
\end{table}

\begin{table}[ht] 
\begin{center}
 \caption{Precision and recall of OntoSeer for use case 2 in the exact setting.}
\label{tab:Table20}
\begin{tabular}{ ccccccc } 
 \hline
Feature&Precision@3 & Recall@3 & Precision@5 & Recall@5 & Precison@7 & Recall@7\\
 
 \hline\hline
 ODP&0.66&0.4 &0.8&0.8&0.571 &0.8\\ 
 Vocabularies&0.66&0.4&0.8&0.8&0.71&1\\ 
 Axioms&0.66&0.33&0.6&0.5&0.71&0.83\\
 \hline

\end{tabular}
\end{center}
\end{table}
\section{Related Work}

Ontology development process involves a series of tasks that are iterative in nature and there is no fixed order for most of the tasks~\cite{article23}. Although there is no one \emph{correct} way of building an ontology, the broad tasks involved in ontology development are as follows. 

\begin{itemize}
    \item List a few competency questions to restrict the scope of the ontology.
    \item Gather the most important terms in the domain.
    \item Split the terms into nouns and verbs. Nouns generally become the classes and verbs become the properties.
    \item Carefully go over the classes and build a hierarchy. The more general classes can be considered as superclasses and the more specific ones become the subclass.
    \item Reuse existing ontologies as much as possible.
    \item The relationship between the concepts should be captured using the axioms.
    \item Appropriate object and data properties should be associated with the classes.
    \item Check whether the ontology can answer the competency questions and if it does not, revise the ontology and repeat some of the listed steps.
\end{itemize}

Each of the above listed steps is subjective and can lead to confusion among the ontology developers. Having the support of a tool in handling these tasks would immensely help the ontology developers. Apart from OntoSeer, there have been a few efforts in this direction. 

Tools that are similar to OntoSeer are  OntoCheck~\cite{DBLP:journals/biomedsem/SchoberTSB12}, and OOPS!~\cite{poveda-villalon_oops!_2014}. We briefly discuss the functionality of these tools and how they differ from OntoSeer.
OntoCheck~\cite{DBLP:journals/biomedsem/SchoberTSB12} is a plugin for Protégé that only indicates the class names that violate the naming convention. For example, if there are classes such as \texttt{HumanBeing}, and  \texttt{nitrogenoxide}, then OntoCheck reports that 50\% of the classes do not follow naming conventions along with the class \texttt{nitrogenoxide}. In contrast, OntoSeer indicates the violation and recommends possible class names such as \texttt{NitrogenOxide} to the user.

OOPS! stands for Ontology Pitfall Scanner. It is an online tool that detects pitfalls or common modelling errors that ontology developers make. The pitfalls are divided into three categories - structural, functional, and usability-profiling. There are around 40 pitfalls, but here we discuss only some of them due to lack of space. Having cycles in the class hierarchy is a pitfall that comes under the structural dimension. Generally, the ontology should not contain more than one semantically equivalent class. Missing disjoint axioms is a common pitfall. Unless explicitly stated, classes are not disjoint. A functional pitfall is connecting disconnected components. Sometimes two unrelated entities are joined by a relationship. This is a pitfall that should be avoided. A usability pitfall is using different naming conventions in the ontology. For example, a superclass will have a name following one naming convention, and a subclass name follows a different naming convention. Missing domain and range for properties is another common pitfall. 

OntoClean~\cite{guarino_overview_2009} provides guidelines for properties and class hierarchies. It introduced terms such as essence, rigidity, identity, and unity. A property that holds in all possible cases is termed as essential for the entity. Rigidity is a type of essential property that is true for all the instances of an entity. A rigid class is a class whose every member cannot cease to be a member without losing its existence. A rigid class cannot be a subclass of an anti-rigid one. Identity is defined as the ability to differentiate two individual entities as different or recognize them as the same. Identity helps in resolving confusion regarding class hierarchies. Unity determines whether an instance is the whole or part of an entity. ODEClean plug-in of WebODE~\cite{DBLP:conf/eon/Fernandez-Lopez02} devised a way to validate class hierarchy with OntoClean taxonomy. Unlike OntoSeer, ODEClean is a stand-alone web plugin.  Also, ODEClean assumes that the user will know the characteristics of each property which may not be true in the case of novice developers. OntoSeer, on the other hand, asks the user simple questions, and based on the answers, validates the class hierarchy.

One of the main differences between OntoClean, OOPS! and OntoSeer is that OntoSeer works along with the ontology developer in creating better quality ontologies. It does not evaluate ontologies post creation. Another major difference is that OntoSeer recommends terms (classes, properties, instances) that can be reused. It does this by checking the similarity of the terms from the ontology being built with existing ontologies. 

An important functionality of OntoSeer is to encourage reusability by making use of existing ontologies and ODPs.  An ODP helps domain experts in ontology engineering by packaging reusable best practices into small blocks of ontology functionality~\cite{DBLP:conf/semweb/Hammar12}. Since there are several hundred ODPs, it is difficult to select the relevant ones. OntoSeer helps users in the selection of ODPs by making use of the classes and properties of the ontology under construction and asking a  few simple questions to the user. This feature is one of major differences between existing tools (described earlier) and OntoSeer.

\section{Sustainability Plan}
We collected as many ontology corpora as possible for indexing and recommendation. But the index that OntoSeer uses may become stale over a period of time. Users can add more ontologies or update the existing corpora. OntoSeer can regenerate the indexes and use them for various recommendations that it supports. Since we query LOV and BioPortal over the Web, the updated ontologies in these portals will get reflected in the recommendations as well. For the ODP recommendation, OntoSeer expects each ODP (with all the details such as description, CQs, axioms, etc.) to be in a separate text file. So the new additions to the ODP repository need to be made available as text files. There is no danger of naming recommendation and class hierarchy validation going stale. Based on the user response and the initial interest shown by the community on the code repository (5 users have \emph{starred} the repository), we expect an increase in the uptake of this tool by the community and this in turn will help with the sustainability of OntoSeer.


\section{Conclusion and Future Work}

Developing an ontology is often confusing and time consuming, especially for the inexperienced developers. They have many choices to make, such as the classes and properties to create or reuse, ODPs to use, the different axioms to include in the ontology and the naming conventions to follow. We developed a Prot\'eg\'e plugin named OntoSeer to ease the process of building an ontology. Our tool recommends classes, properties, axioms, and ontology design patterns in real-time while a user is developing the ontology. We have conducted a user study of the tool. There were 21 respondents. Almost all the users are satisfied with the recommendations provided by OntoSeer. The majority of the users (14 out of 21) said that OntoSeer also helps them in saving modelling time. In our evaluation, we also show that the recommendations generated by OntoSeer are relevant and are of good quality. In the future, we plan to engage in a dialogue with the ontology developer in real-time to resolve confusing issues such as a term being a class vs. a property vs. an instance. A dialogue with the user also helps in checking whether the ontology can answer the competency questions.

\textbf{Acknowledgement.} This work has partially been supported by the Infosys Centre for Artificial Intelligence (CAI), IIIT-Delhi, India.

\bibliographystyle{splncs04}
\bibliography{ontquality}

\end{document}